\documentclass{article}

\usepackage{microtype}
\usepackage{graphicx}
\usepackage{subfigure}
\usepackage{tabularx}
\usepackage{booktabs} 

\usepackage{hyperref}
\usepackage{algpseudocode}
\usepackage{multirow}

\usepackage[accepted]{icml2025}

\usepackage{amsmath}
\usepackage{amssymb}
\usepackage{mathtools}
\usepackage{amsthm}

\usepackage{enumitem}
\setlist[itemize]{itemsep=2pt, parsep=0pt, topsep=0pt}

\usepackage[capitalize,noabbrev]{cleveref}

\theoremstyle{plain}

\theoremstyle{definition}

\theoremstyle{remark}

\usepackage[textsize=tiny]{todonotes}

\newcommand{\name}{\textit{Atlas}}

\icmltitlerunning{Multi-Scale Attention Improves Long Context Image Modeling}

\begin{document}

\twocolumn[
\icmltitle{\name: Multi-Scale Attention Improves Long Context Image Modeling}



\icmlsetsymbol{equal}{*}
\icmlsetsymbol{lead}{$\dagger$}

\begin{icmlauthorlist}
\icmlauthor{Kumar Krishna Agrawal}{equal,ucb,lead}
\icmlauthor{Long Lian}{equal,ucb}
\icmlauthor{Longchao Liu}{ucb}
\icmlauthor{Natalia Harguindeguy}{ucb,ucsf}
\icmlauthor{Boyi Li}{ucb}
\icmlauthor{Alexander Bick}{vdb}
\icmlauthor{Maggie Chung}{ucsf}
\icmlauthor{Trevor Darrell}{ucb}
\icmlauthor{Adam Yala}{ucb,ucsf}
\end{icmlauthorlist}

\icmlaffiliation{ucb}{University of California, Berkeley}
\icmlaffiliation{vdb}{Vanderbilt University }
\icmlaffiliation{ucsf}{University of California San Francisco}

\icmlcorrespondingauthor{Kumar Krishna Agrawal}{kagrawal@berkeley.edu}
\icmlcorrespondingauthor{Adam Yala}{yala@berkeley.edu}

\icmlkeywords{Machine Learning, ICML}

\vskip 0.3in
]



\printAffiliationsAndNotice{\icmlEqualContribution} 

\begin{abstract}
Efficiently modeling massive images is a long-standing challenge in machine learning.
To this end, we introduce Multi-Scale Attention (MSA). 
MSA relies on two key ideas, (i) multi-scale representations (ii) bi-directional cross-scale communication. MSA creates $\mathcal{O}(log N)$ scales to represent the image across progressively coarser features and leverages cross-attention to propagate information across scales. We then introduce \name{}, a novel neural network architecture based on MSA. 
We demonstrate that \name{} significantly improves the compute-performance tradeoff of long-context image modeling in a high-resolution variant of ImageNet 100.  At 1024px resolution, Atlas-B achieves 91.04\% accuracy, comparable to ConvNext-B (91.92\%) while being 4.3x faster. Atlas is 2.95x faster and 7.38\% better than FasterViT, 2.25x faster and 4.96\% better than LongViT. In comparisons against MambaVision-S, we find Atlas-S achieves 5\%, 16\% and 32\% higher accuracy at 1024px, 2048px and 4096px respectively, while obtaining similar runtimes. Code for reproducing our experiments and pretrained models is available at \url{https://github.com/yalalab/atlas}.
\end{abstract}

\section{Introduction}
\label{sec:introduction}

\begin{figure}[t]  
    \centering

    \includegraphics[width=0.9\linewidth]{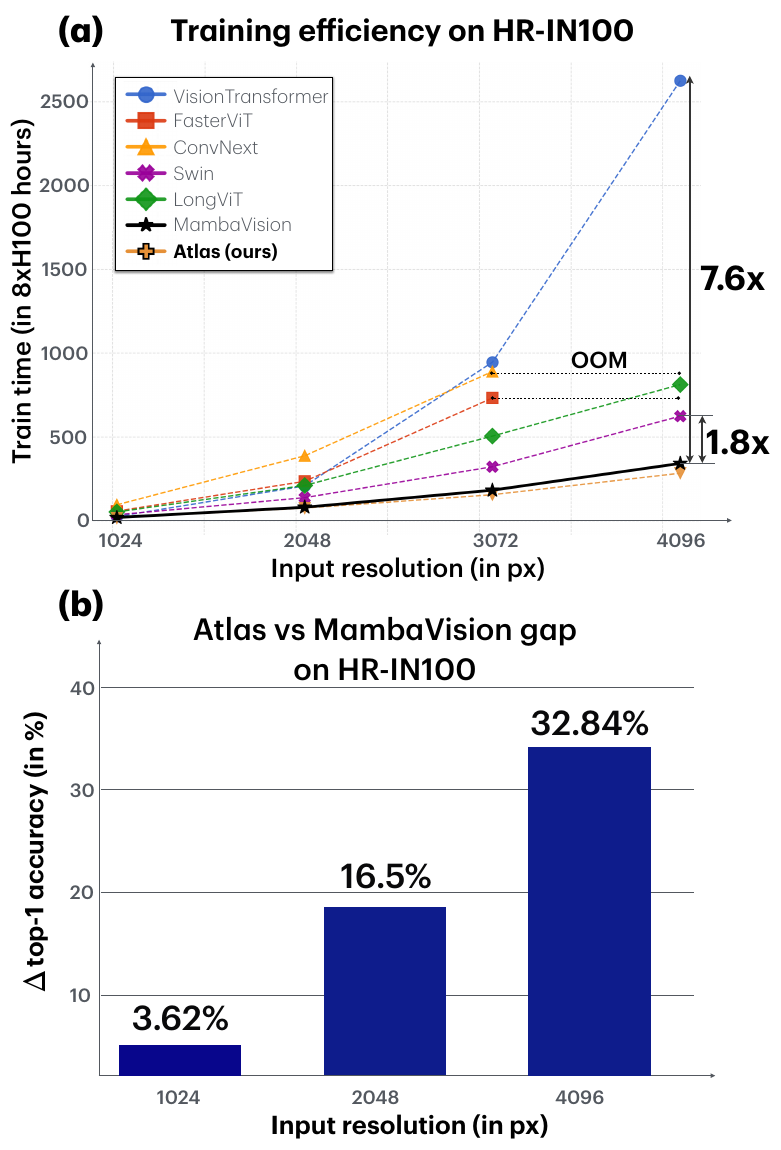}
    \caption{(a) Training efficiency comparison of different vision architectures on HR-IN100 across increasing input resolutions (1024-4096px). (b) \name{} exhibits similar runtime scaling as MambaVision while obtaining significantly better accuracy.}
    \label{fig:training_efficiency}
\end{figure}

Long-context image modeling remains a fundamental challenge in computer vision with broad applications to biomedicine~\cite{xu2024comphistvit}, satellite imagery~\cite{rad2024wacv}, and vision-language modeling~\cite{team2023gemini,Qwen2VL,qwen2.5-VL,chen2024expanding}.
At the core of this challenge is a compute expressivity trade-off; we aim to develop models that efficiently scale to massive input sequences while capturing arbitrary pair-wise dependencies between input tokens. As shown in \cref{fig:training_efficiency}(a), self-attention, as used in Vision Transformers, is highly expressive, but its computational cost scales poorly (i.e., quadratically) with sequence length. It remains infeasible to train end-to-end Vision Transformers on massive imaging modalities such as mammograms or whole-slide pathology images.
At another end of the spectrum, state space models (SSMs) and recurrent architectures are highly efficient, achieving linear computational complexity; however, SSM-based models perform poorly in long-context imaging modeling (\cref{fig:training_efficiency}b).

Long-context image modeling requires novel neural primitives and new benchmarks to guide their development. Recent work in efficient architecture design, such as FasterViT \cite{hatamizadeh2023fastervit} and MambaVision \cite{hatamizadeh2024mambavision}, has primarily focused on improving the throughput vs accuracy trade-offs in the context of standard resolution ImageNet experiments ($224\times224$). 
While valuable, this setting yields little insight into how methods scale to larger input resolutions. 
To this end, we propose a new high-resolution benchmark based on ImageNet-100 (HR-IN 100). 
We evaluate the speed vs accuracy trade-off of different neural networks at progressively larger resolutions, ranging from $1024\times1024$ to $4096\times4096$ images. 
As input resolution increases, long-range communication across distant parts of the image becomes more essential for image classification, and asymptotic computational complexity begins to dominate model runtime. 

In designing a novel neural primitive, we aim to enable arbitrary cross-token interaction with minimal intermediate steps (i.e., communication complexity) while minimizing computational complexity as a function of input sequence length. 
To this end, we propose Multiscale Attention (MSA), a novel primitive for high-resolution image modeling. 
MSA is built on two key ideas: multiscale representations and cross-scale communication. 
In each MSA block, we leverage a simple $S$-token max-pooling kernel to summarize small spatial regions (e.g., 4x4 input region), into progressively coarser summary representations across $O(\log_{S}N)$ spatial scales, where $N$ is the total sequence length.
We then leverage a windowed cross-attention mechanism to enable information-sharing between tokens at different scales.
At each scale,  tokens attend to nearby tokens of the same scale and tokens from all coarser scales.
This ``top-down'' communication enables MSA to integrate information across the entire sequence.
Each scale's tokens also cross-attend to its ``parent'' finer-grain scale tokens, allowing each coarse token to refine its representation through ``bottom-up'' communication. 
Altogether, this bi-directional communication pattern enables information mixing between all input tokens through $O(\log N)$ intermediate tokens (i.e. coarse scale representations) and within $O(N \log N)$ runtime. 
In controlled block-level experiments (see \cref{tab:block-comparison}), we find that MSA outperforms alternative neural network primitives in long-context modeling,  including LongNet's dilated attention \cite{ding2023longnet}, MambaVision Mixer \cite{hatamizadeh2024mambavision}, and FasterViT's Hierarchical Attention \cite{hatamizadeh2023fastervit}.

We propose \name{}, a novel architecture designed around the unique advantages of MSA. Given a sequence length N, which defines $\log N$ scales within MSA, Atlas leverages $\log N$ macro-stages to progressively down-sample the input until MSA recovers only a single scale. We leverage the rich scale-2 representations of our MSA block as a down-sampling mechanism, enabling both faster and more performant down-sampling than traditional approaches. We demonstrate that \name{}  significantly improves the Pareto frontier in long-context modeling.  In $1024\times1024$ experiments, as shown in \cref{tab:model-comparison}, \name{} obtains comparable runtime to MambaVision (23.1hr vs 22.6hr) on the same hardware, while obtaining 6.1\% higher accuracy (91.04 vs 84.86). Compared to FasterViT and LongViT, \name{} is $2.95 \times$ and $2.25 \times$ faster, obtaining 7.38\% (91.04 vs 83.66) and 4.96\% (91.04 vs 86.08) higher accuracy, respectively. Moreover, the performance advantage of \name{} is especially pronounced as we scale input resolution to 4096px, achieving a 34\% accuracy improvement over MambaVision at similar runtime.

We summarize our contributions as follows:\vspace{-2mm}
\begin{itemize}
    \item We propose a High-Res ImageNet-100 (HR-IN 100), an efficient benchmark with input resolutions ranging from $1024\times1024$ to $4096\times4096$ for evaluating the frontier of long-context image modeling.
    \item We introduce Multi-Scale Attention (MSA), a novel neural network primitive that maintains representations across $O( \log N)$ spatial scales and enables bi-directional information mixing across all scales within $O(N \log N)$ runtime. Building on MSA, we introduce \name{}, a novel neural network architecture.
    \item With extensive experiments on High-Res ImageNet-100, we demonstrate that \name{} improves the Pareto frontier in long-context image modeling. \name{} outperforms representative efficient architectures in long-context image modeling, including FasterViT, MambaVision, and LongViT.
\end{itemize}

\begin{figure*}[t]
    \centering
    \includegraphics[width=\textwidth]{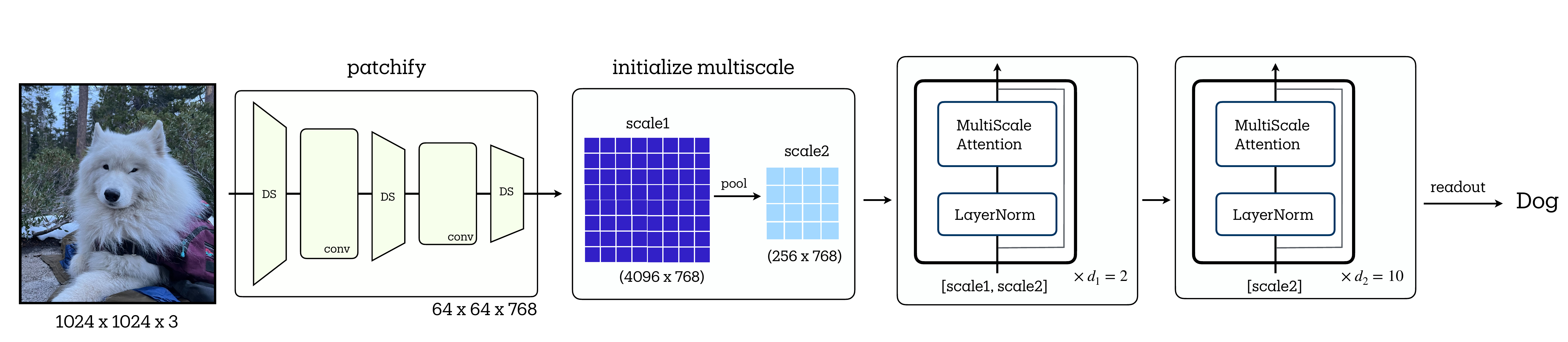}
    \vspace{-5mm}
    \caption{\textbf{The Atlas architecture} consists of a convolutional stem for initial feature extraction, followed by a series of Multi-Scale Attention (MSA) blocks that progressively downsample the feature maps while preserving global context. This hierarchical design facilitates the effective processing of high-resolution images with efficient communication between features.\vspace{-2mm}}
    \label{fig:msa}
\end{figure*}

\section{Related Work}
\label{sec:related_work}
\noindent\textbf{Vision Transformers (ViTs).} 
ViTs \cite{dosovitskiy2020image} directly apply Transformers \cite{vaswani2017attention} architecture to image patches, demonstrating the effectiveness of self-attention in visual data processing. Building on this,  DeiT \cite{touvron2021training} improves training data efficiency. However, the self-attention primitive in ViT, which scales quadratically with input sequence length, limits its application toward high-resolution imaging. Our study focuses on developing efficient alternatives to standard self-attention to enable expressive and computationally efficient long-context image modeling.

\noindent\textbf{Efficient Long Sequence Modeling in Language.} 
To address the challenges of long-sequence modeling, LongNet \cite{ding2023longnet} introduces a dilated attention mechanism, allowing transformers to process sequences with up to one million tokens.  LongNet was later adapted into a vision model as LongViT \cite{wang2023image} to process whole-slide pathology images. State Space Models (SSMs), such as Mamba \cite{gu2023mamba}, provide a linear-time alternative to full attention for efficient long sequence modeling. RetNet \cite{sun2023retentive} combines the strengths of recurrence and attention, enabling linear-time sequence modeling. Longformer \cite{beltagy2020longformer} integrated local and global attention for effective long-document processing. Our work is most similar to LongNet, which also achieves a communication complexity of $\mathcal{O}(\log N)$, where $N$ is the length of the input sequence.

Instead of using dilated attention, we propose multiscale attention (MSA), which captures distant dependencies by attending to a subset of the input through intermediate ``coarser scale'' tokens. Unlike dilated attention, MSA effectively leverages locality in the input, resulting in significantly improved long-context vision modeling.

\noindent\textbf{Efficient Visual Modeling.} 
Vim and VMamba \cite{zhu2024vision,liu2024vmambavisualstatespace} adapted  State-space models (SSMs) to vision-specific tasks and demonstrated the effectiveness of SSMs for visual representation learning. MambaVision \cite{hatamizadeh2024mambavision} proposed a hybrid SSM and self-attention-based architecture, and demonstrated improved performance over other SSM-based architectures.
Swin \cite{liu2021swin} proposes leveraging window-shifting for cross-window communication and a hierarchical design to aggregate context. CSwin \cite{dong2022cswin} proposes cross-shaped window attention to capture global and local dependencies. CrossViT \cite{chen2021crossvit} uses a dual-branch architecture to process image patches of varying sizes.
EdgeViT \cite{pan2022edgevits} and EfficientFormer \cite{li2022efficientformer} designed lightweight transformers that are specially optimized for edge devices. VisFormer \cite{chen2021visformer} combined convolutions and transformers for vision tasks. Twins \cite{chu2021twins} improved the spatial attention mechanisms for improved performance. The long-short transformer \cite{zhu2021long} introduced hybrid attention for efficient modeling in vision and language. FasterViT \cite{hatamizadeh2023fastervit} introduced hierarchical attention for fast visual information processing, and demonstrated improved performance over Swin, Twins, CrossViT, and EfficientFormer. Focal Transformer \cite{yang2021focal} explores a new form of attention and Pyramid Vision Transformer (PVT) \cite{wang2021pyramid} explores hierarchical attention for efficient modeling. Unlike these works, which focus on improving compute-accuracy trade-offs in modest resolution regimes (i.e. 224 x 224 pixels), our work focuses on modeling high-resolution images. In this context, we find that representative efficient architectures, including MambaVision and FasterViT, fail to effectively process high-resolution images.  

\noindent\textbf{Multi-resolution representations in Neural Networks.}
Dense cross-scale communication has been explored in the context of CNNs, as in DenseNets~\cite{huang2017densely} and Feature Pyramid Networks (FPNs)~\cite{lin2017feature}.  In these works, feature maps across multiple resolutions are integrated using fixed operations, including concatenation or summation. In contrast, we propose to fuse representations across scales in our MSA block through cross-attention. This data-dependent multi-scale integration strategy allows our model to learn complex interactions between features at different resolutions and optimize the fusion process jointly with feature extraction.

\section{Method}

We propose Multi-Scale attention (MSA), a novel neural primitive for long-context image modeling. MSA builds representations across multiple spatial scales and leverages dense cross-attention operations to share information across scales. Building on this primitive, we build \name , a hierarchical macro-architecture that uses the intermediate scales in MSA as a novel down-sampling mechanism.

\begin{figure*}[t]
    \centering
    \includegraphics[width=0.9\textwidth]{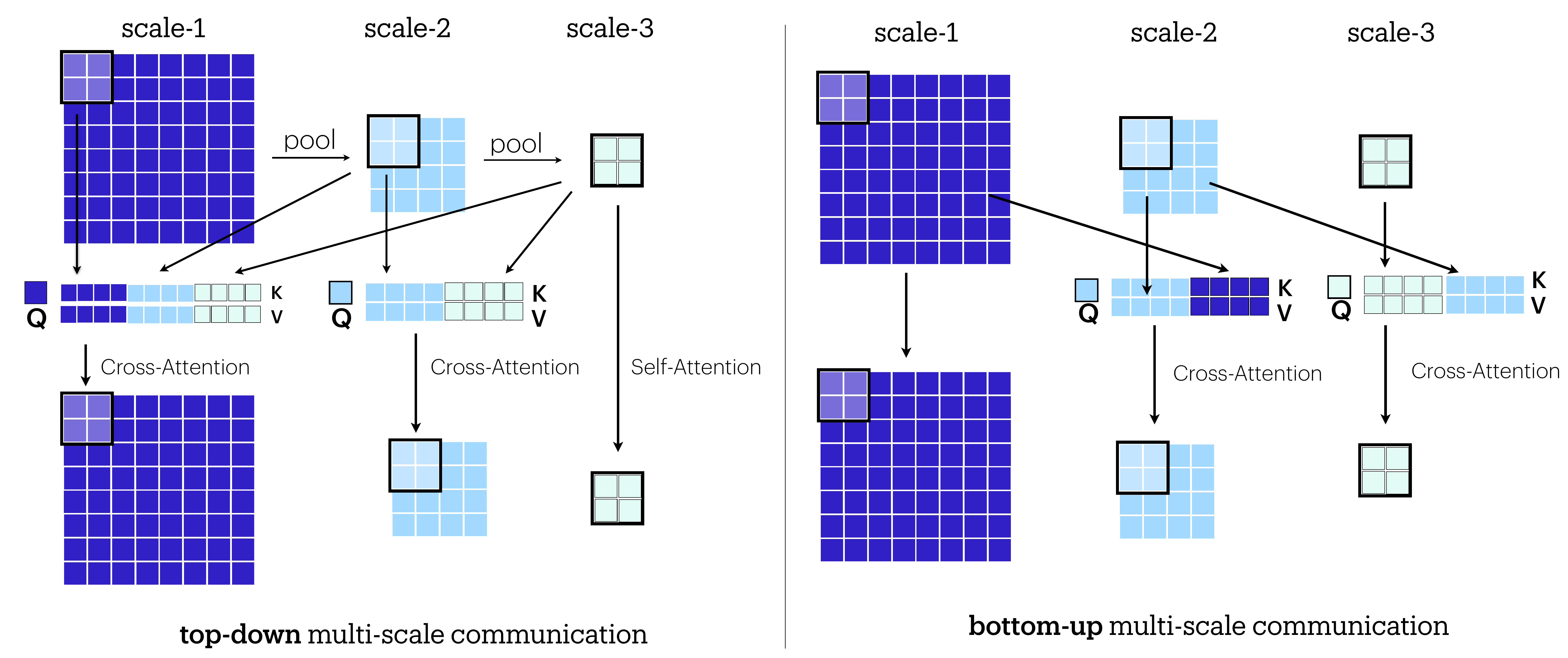}
    \vspace{-5mm}
    \caption{\textbf{Illustration of top-down and bottom-up hierarchical communication} in Multi-Scale Attention (MSA). The top-down Global Context Aggregation enables coarse-to-fine feature propagation. The bottom-up fine-to-coarse pathway propagates high resolution features into coarser scale representations. }
    \label{fig:msa_topdown}
\end{figure*}

\subsection{Preliminaries}

Windowed self-attention (WA) adapts the standard Multi-Head self-attention (MHSA) to operate efficiently on local regions of an input feature map. To lay the groundwork for multi-scale attention (MSA), we first describe the WA operation and analyze its computational benefits and limitations.

\noindent\textbf{Windowed Self-attention.} Consider a feature map $X \in \mathbb{R}^{H \times H \times C}$, where $H$ is the spatial and $C$ is the channel dimension\footnote{For simplicity, we focus on square 2D feature maps, but the concept is generalizable to 1D sequences or 3D volumes.}. The WA mechanism operates in two key steps:

1. \textbf{Window Partitioning}: Divide the feature map into non-overlapping windows of size $k \times k$, with the number of windows per dimension: $H' = H/k$, total number of windows $M = H' \times H' = (H/k)^2$, and each window $W_{ij}$ ($i,j \in \{1,\ldots,H'\}$) containing $k^2$ tokens. Further, each window $W_{ij}$ is reshaped into a sequence, where $W_{ij} \in \mathbb{R}^{k \times k \times C}$ is viewed as $W_{ij} \in \mathbb{R}^{k^2 \times C}$ after reshape.

2. \textbf{Local Self-Attention}: Apply standard Multi-Head Self-Attention (MHSA) within each window:
\begin{align*}
Attention(Q, K, V) &= \text{softmax}\left(\frac{Q K^T}{\sqrt{d_k}}\right)V \\
\text{where } Q,K,V &= \text{Linear Projections}(W_{ij})
\end{align*}

The computational complexity of WA within a single window is $O(k^2 \cdot k^2) = O(k^4)$ due to the attention operation within $k^2$ tokens.  Since there are $M = (H/k) \times (H/k) = H^2/k^2$ windows, the total complexity of WA across the entire feature map becomes $O(M \cdot k^4) = O(\frac{H^2}{k^2} \cdot k^4) = O(H^2 k^2) = O(N k^2)$, where $N = H^2$ is the total number of tokens in the feature map.  This is a significant reduction from the $O(N^2)$ complexity of global attention, especially when $k \ll \sqrt{N}$.

While computationally efficient, WA suffers from two key limitations: 1) \textbf{Limited Receptive Field}: Each window processes information independently, preventing direct communication between different image regions, and 2) \textbf{Boundary Effects}: Objects or features spanning multiple windows cannot be directly modeled within a single attention operation. For example, an object split across two windows must be processed independently in each window, relationships between parts can only be learned indirectly when all features merged at the final readout.

\begin{figure}[b]
    \centering
    \includegraphics[width=1.0\linewidth]{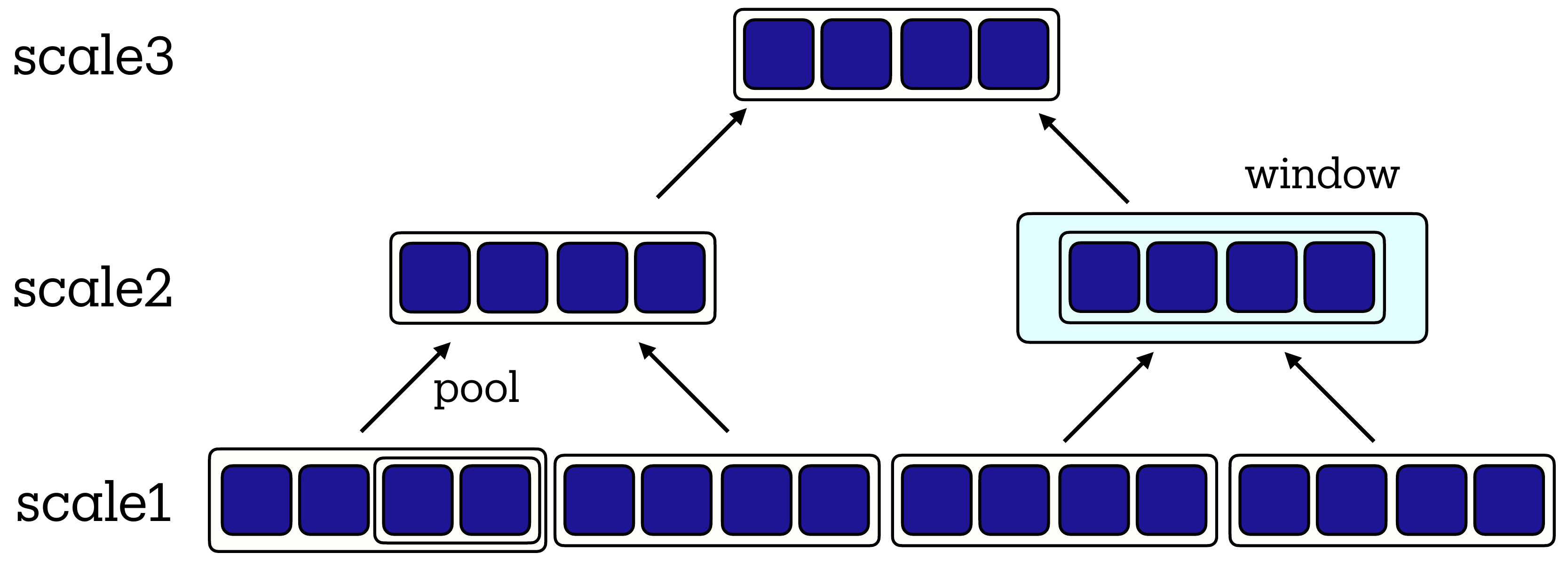}
    \vspace{-5mm}
    \caption{\textbf{Multi-Scale features} with iterative summarization.}
    \label{fig:multiscale}
\end{figure}

\begin{figure*}[t]
    \begin{minipage}{.49\linewidth}
      \begin{algorithm}[H]
        \small
        \algrenewcommand\algorithmicrequire{\textbf{Input: }}
        \algrenewcommand\algorithmicensure{\textbf{Output: }}
        \caption{Multi-Scale Attention (MSA) Block}
        \label{alg:msa_block}
        \begin{algorithmic}[1]
          \Require $\mathcal{X} = [X^{(1)}, ..., X^{(L)}]$, where  $X^{(l)} : \mathtt{(B, N_l, C_{in})}$ \newline $k\times k \gets \text{Window Size}$, $S \gets \text{Downsampling Rate}$
          \Ensure $\mathcal{\overline{X}} = [X^{(1)}, ..., X^{(L)}]$, where  $X^{(l)} : \mathtt{(B, N_l, C_{in})}$
          \newline
          \newline \Comment{\textit{Iterative summarization}}
          \For{$l = 2, ..., L$} \Comment{Iterate from fine to coarse}
              \State $X^{(l)} \mathrel{{+}{=}} \mathsf{Summarize}(X^{(l-1)}, S)$ \Comment{\cref{eq:summarize}}
          \EndFor \newline \newline
          \Comment{\textit{Top-Down Communication: Global Context Aggregation}}
          \For{$l = L, L-1, ..., 1$} \Comment{Iterate from coarse to fine}
                  \State $X^{(l)} \gets \mathsf{CrossAttention}(X^{(l)}, [X^{(l)}, X^{(l+1)}, ..., X^{(L)}])$ \newline \phantom{xxx}  \Comment{as in \cref{eq:topdown}}
          \EndFor
          \newline \newline\Comment{\textit{Bottom-Up Communication: Fine-to-Coarse Refinement}}
          \For{$l = 2, 3, ..., L$} \Comment{Iterate from fine to coarse}
              \State $X^{(l)} \gets \mathsf{CrossAttention}(X^{(l)}, X^{(l-1)})$
              \newline \phantom{xxx}  \Comment{as in \cref{eq:bottomup}}
          \EndFor
          \State \textbf{return} $\mathcal{\overline{X}}  = [X^{(1)}, ..., X^{(L)}]$
        \end{algorithmic}
      \end{algorithm}
    \end{minipage}%
    \hfill%
    \begin{minipage}{.49\linewidth}
      \begin{algorithm}[H]
        \small
        \algrenewcommand\algorithmicrequire{\textbf{Input: }}
        \algrenewcommand\algorithmicensure{\textbf{Output: }}
        \caption{Atlas Architecture Pseudocode}
        \label{alg:atlas_icml}
        \begin{algorithmic}[1]
          \Require $\text{Img} : \mathtt{(B, H_{in}, W_{in}, C_{in})}$, $k \times k \gets \text{window size}$ \newline $P \gets \text{Patch Size}$, $S \gets \text{Downsampling Rate}$ \newline $D \gets \{d_1, d_2, ..., d_L\}$ \Comment{Atlas Configuration}
          \Ensure $\text{predictions} : \mathtt{(B, D_{out})}$ \Comment{downstream predictions} \newline
  
          \State $X^{(1)} \gets \mathsf{ConvPatchify}(\text{Img}, P)$ \Comment{scale 1 feature map} \newline
          \newline \Comment{\textit{Initialize Multi-Scale features}}
          \For{$l = 2, ..., L$}
              \State $X^{(l)} \gets \mathsf{Summarize}(X^{(l-1)}, S)$ \Comment{Strided MaxPool}
          \EndFor
          \newline
          \newline \Comment{\textit{Progressive Downsampling}}
          \For{$s = 1, 2, ..., L$} \Comment{Iterate through stages}
              \For{$blk = 1, 2, ..., d_s$}  
                  \State $[X^{(s)}, ..., X^{(L)}]$$ \gets \mathsf{MSABlock}([X^{(s)}, ..., X^{(L)}]$, k, S)
                  \State \Comment{Apply MSA Block}
              \EndFor
          \EndFor \newline
          \State $\text{predictions} \gets \mathsf{readout}(X^{(L)})$
          \State \textbf{return} $\text{predictions}$
        \end{algorithmic}
      \end{algorithm}
    \end{minipage}
  \end{figure*}

\subsection{Multi-Scale Attention}

MSA's core design centers on two key components: 1) a \textbf{hierarchical representation} that creates intermediate feature scales using fixed-size summarization kernels to preserve information density and 2) \textbf{bi-directional communication} that enables effective information exchange across multiple windows and scales, through dense cross-attention.

\subsubsection{Hierarchical Representation}
\label{sec:latents}
Multi-Scale Attention (MSA) builds hierarchical representations through iterative summarization with a fixed-size kernel of $S$-tokens. Starting with the input feature map $F^{(1)}$ at scale-$1$, we create coarser representations through a summarization operation $\mathcal{S}$:
\begin{align}
    F^{(l)} = \mathcal{S}(F^{(l-1)}, S), \quad \text{for } l = 2, \ldots, L
    \label{eq:summarize}
\end{align}
where $\mathcal{S}$ is implemented as strided max-pooling with a fixed stride $s$ (i.e. downsampling rate $S=s \times s$ tokens). This process continues until the feature map size at scale $L$ is no larger than the window size $k \times k$. With input sequence length $N$ and downsampling rate $S$, the number of scales $L$ grows logarithmically as $O(\log_S{N})$, where $S=s^2$.

At each scale $l$, we operate on windows, by partitioning the feature map $F^{(l)}$ into non-overlapping regions of size $k \times k$ (i.e. $K=k^2$ tokens), yielding windows $\{W^{(l)}_{ij}\}$, for $l=1, .., L$ scales. As shown in \cref{fig:multiscale},  this scheme creates a directed acyclic graph (DAG) between windows at different spatial scales. With every summarization operation, we merge “parent” windows into new coarser “child” windows. 

\subsubsection{Cross-Scale Communication: Attention-Based Fusion}
The expressive power of MSA comes from its ability to efficiently propagate information across scales through two complementary mechanisms:

\textbf{I. Top-Down Communication}

In our top-down communication scheme, we propagate information from coarse ``child'' windows to their ``parent'' windows through a dense set of cross-attention operations.  

Let $W^{(l)}$ be a window at scale $l$, and $\{W^{l+1}, \ldots, W^{L}\}$ denote the corresponding coarse "child" windows as illustrated in \cref{fig:multiscale}. The cross-attention operation using standard Multi-Head Attention (MHA), as visualized in \cref{fig:msa_topdown}, is then given by:
\begin{align}
    W^{(l)} = \text{MHA}(Q_l, [K_{l:L}], [V_{l:L}])
    \label{eq:topdown}
\end{align}
where $Q_l, K_l, V_l$ are query/key/value projections of $W^{(l)}$, and $K_{l+1:L}, V_{l+1:L}$ are concatenated key/value projections from coarser scales. This operation enables MSA to model relationships between tokens within the window, while also allowing each window to {\it read} from long-context information from all coarser scale "child" windows. This dense cross-attention design allows each scale to directly observe global context through the coarsest scale "child" window. At the coarsest scale $l=L$, this operation recovers standard self-attention.

\textbf{II. Bottom-Up Communication}

The bottom-up communication in MSA complements the top-down aggregation by refining coarser-scale "child" representations with detailed information from finer-grain "parent" tokens. This is a localized operation, in the sense that the fine grain refinement for each token is guided only by its {\it direct} parent window.

Specifically, let $Q_{l}$ be the query projection of $W^{l}$, and let $K_{l-1}$ and $V_{l-1}$ be the key and value projections from the parent window $W^{(l-1)}$. The updated window representation $W^{(l)}$ after bottom-up communication is obtained through cross-attention as:
\begin{align}
    W^{(l)} &= \text{MHA}(Q_{l}, K_{l-1}, V_{l-1})
    \label{eq:bottomup}
\end{align}
This targeted cross-attention allows for the recovery and integration of crucial local information potentially lost in the initial summarization.

The pseudocode for the full MSA block is shown in \cref{alg:msa_block}.

\noindent{}\textbf{Asymptotic Complexity.} With a feature map $X \in \mathbb{R}^{N \times C}$ and window of $K$ tokens (typically $K= k\times k$), downsampling rate $S$, MSA creates $L=\log_S{N}$ feature scales. The most expensive operation is the dense \textit{top-down} cross-attention. In particular, for scale-1, each token cross attends to $LK$ tokens (one window per scale), which scales to $NLK$ complexity across a $N$-length sequence. The runtime for all subsequent scales $2, .., L$ is upper-bounded by $NLK$, giving an effective runtime complexity of $O(NLK)$.

Plugging in $L=\log_S{N}$, we recover $\mathcal{O}(NK\log_S{N})$ as the net runtime complexity of \name{}. Note that $K$ and $S$ are typically small constants depending on the hardware; in our experiments we find $K=256$ (i.e. $16\times 16$ window) and $S=16$ (i.e. $4\times4$) to be most performant on an $8\times$H100 node. Our dense cross-scale communication strategy guarantees that each token must propagate information across at most $\mathcal{O}(\log_S{N})$ intermediate tokens to interact with any another token in the sequence, where standard self-attention would obtain $\mathcal{O}(1)$ communication complexity but $O(N^2)$ runtime complexity.

\newcommand{\arch}{{Atlas}}

\subsection{ \name{} }
\label{sec:atlas}

The MSA block can be used as drop-in replacement for the standard MHA block in existing architectures like ViT \cite{dosovitskiy2020image} or Swin Transformer \cite{liu2021swin}. To fully leverage the benefits of MSA, we co-design the network structure for \name{}  to optimize performance and efficiency. Our full architecture is illustrated in \cref{fig:msa}, with the pseudocode in \cref{alg:atlas_icml}.

Atlas is a multi-stage architecture, with a convolutional stem \cite{hatamizadeh2023fastervit, hatamizadeh2024mambavision, xiao2021early}, followed by multiple stages of MSA blocks. We leverage the same convolutional stem as FasterViT \cite{hatamizadeh2023fastervit} to obtain localized patch-level representations. In particular, the stem utilizes two stages of residual convolutional blocks, yielding in feature map of $\mathbb{R}^{H/16 \times W/16 \times C}$. Given this feature map, fixed window size $K$ and downsampling rate $S$, MSA builds a multi-scale layout with $L = \log_S{N}$ scales, as outlined in \cref{sec:latents}.

As part of the co-design of Atlas, we fix the number of stages of MSA blocks to be identical to the number of scales, i.e. $L = \log_S{N}$. The key insight behind Atlas is to progressively reduce the number of tokens at each scale, focusing computational resources on high-level features. Given the multiscale structure of the MSA block, we propose a progressive scale-dropping strategy in Atlas. In other words, for a multi-scale input $\mathcal{X}=[X^{(1)}, X^{(2)}, ..., X^{(L)}]$, stage $l$ of the \name{} only processes $[X^{(l)}, X^{(l+1)}, ..., X^{(L)}]$ actively.

As a concrete instance, for MSA with $L$ scales, let us define an Atlas config $D=\{d_1, d_2, ..., d_L\}$ with $L$ corresponding stages. Here, $d_l$ is the number of blocks at stage $l$. For example, for a 4-scale MSA block, an Atlas config would have 4 stages, e.g. $D = \{2, 2, 2, 6\}$. This config indicates that the first scale is the finest resolution for the first two blocks, after which it becomes inactive and is dropped. Subsequently, the second scale becomes the finest active resolution for the next two blocks, with $X^{(4)}$ being the only active features for the last block. 

This strategy is quite flexible, in that for a single scale MSA block, and $K=N$, it recovers the standard ViT with MHSA block. For the readout, there are multiple strategies to aggregate the final representations across scales. We find that simply using the last scale as the final representation works well in practice. 

\section{Experiments}
\label{sec:experiments}

\subsection{Image Classification}

\noindent{}\textbf{Setup.} 
We propose using a novel high-resolution benchmark based on Imagenet-100 \cite{tian2020contrastive}, High-Resolution ImageNet-100. The dataset extends the original Imagenet-1k dataset with $\sim$126K unique training samples, 5000 validation samples, and 100 classes with high-resolution images (up to 4096px), where the images are upsampled to the desired resolution. 
We first focus on a system's level comparison against representative architectures, including ViT, Swin, FasterViT, MambaVision, ConvNext, and LongViT. Together, these architectures encompass sparse attention, SSMs, convolutional, and dilated attention approaches. 
For each baseline , we utilize the provided code as is, without modifications to gradient accumulation, employing a linearly decaying learning rate proportional to the batch size, following~\cite{goyal2017accurate}. This ensures consistency with prior work and facilitates a fair comparison.

{\bf Comparing Architectures}: We benchmark all architectures on the same hardware, 1 server with 8$\times$H100 Nvidia GPUs using 1024px input resolution (equivalent to 4K tokens with patch-size=16). To understand the runtime-performance tradeoff of \name{} design against existing architectures, we train Base-scale models (i.e. 12 head, 768 embed-dim
following prior work \cite{dosovitskiy2020image}) for 320 epochs. 

{\bf Long-Context Image Modeling}: To understand the efficacy of \name{} in long-context image modeling tasks, we seek to scale the evaluation to higher resolutions. Due to extreme cost of running our baselines for full convergence runs (320 epochs) at Base models, we focus our scaling experiments on only our two fastest models, namely Atlas and MambaVision in Small regime. As shown in \cref{fig:training_efficiency}, all other architectures are significantly slower at higher resolutions. Prior work in architecture design for vision models \cite{xiao2021early} demonstrates meaningful comparisons with shorter training schedules. We adopt a similar approach and train \name{}-S and MambaVision-S models for 100 epochs for 1024px, 2048px and 4096px, scaling upto 64K tokens.

\subsection{Ablations}

\noindent{}\textbf{Attention Mechanism.} To understand the efficacy of different token-mixing (e.g. attention or SSM-based) blocks in long-context image modeling, we conduct controlled experiments, using the same optimizer, learning rate schedules. We consider $384\times 384$ inputs, with $4\times 4$ patches, giving a sequence length $N=9216$. We use 4-block architectures, with Base-equivalent blocks (i.e. 12 head, 768 embed-dim following prior work \cite{dosovitskiy2020image}. We compare our MSA block with Hierarchical Attention block \cite{hatamizadeh2023fastervit}, MambaVision Mixer \cite{hatamizadeh2024mambavision}, dilated attention with the LongViT block \cite{ding2023longnet} and standard ViT, Window-ViT blocks.

\noindent{}\textbf{Communication Mechanism.} Our proposed Multi-Scale Attention (MSA) block relies on bi-directional communication to effectively model long-context. To understand the contribution of each mechanism, we conduct controlled ablations with  $256\times 256$ inputs, using $4\times 4$ patches, giving a sequence length $N=4096$, K=256 (i.e. $16\times 16$ windows), S=16 (i.e. $4\times4$ strided max-pool). In this setting we have features at two scales, providing a sandbox to test the contribution of different communication mechanisms. We use a Small-scale 4-block architecture (i.e. with 6 heads, 384 dim  following \cite{dosovitskiy2020image}). The predictions from both scales are merged via average pool, before readout. In this setting, we compare the following variants of the block 
\begin{itemize}
    \item {\bf no-multiscale} : equivalent to vanilla single-scale WA
    \item {\bf no communication}: equivalent to WA at both scales.
    \item {\bf top-down only}:  propagates from coarse to fine only
    \item {\bf bottom-up only}: propagates from fine to coarse only
    \item {\bf top-down + bottom-up}: both mechanisms as in MSA.
\end{itemize}

\noindent{}\textbf{Composition Strategies.} To identify the best MSA block composition strategy, we compare three different strategies of incorporating MSA
\begin{itemize}
    \item {\bf stack}: vanilla stacking of blocks as in \cite{dosovitskiy2020image}, with averaging tokens across scales for readout.
    \item {\bf convolutional downsampling} : similar to prior work as in \cite{liu2021swin,hatamizadeh2023fastervit} we use separate downsampling layer to reduce spatial resolution by $2\times2$ per stage. For this variant, we use a uniform 4-stage architecture, i.e. $\{3, 3, 3, 3\}$
    \item {\bf \name{}}: a $\{d_1$=2, ${d_2}$=10\} config outlined in \cref{sec:atlas}
\end{itemize}
 We run each ablations with  $512\times 512$ inputs, using $8\times 8$ patches, giving a sequence length $N=4096$, with 12 Small-scale MSA blocks.

\section{Results}
\label{sec:results}

\subsection{Image Classification}

{\bf Comparing Architectures at 1024px resolution}: The experimental results in~\cref{tab:model-comparison} demonstrate that Atlas-B/16 is competitive with/outperforms existing vision backbones in accuracy, while being computationally efficient. In particular,  Atlas achieves 91.04\% accuracy while delivering substantial speed advantages: 4.3× faster than ConvNext-B (91.92\%), 1.15× faster than ViT (90.66\%), and 1.6× faster than Swin (90.89\%) with competitive accuracy.  Compared to other sparse-transformer backbones, \name{} is 2.95x faster and 7.3\% better than FasterViT, 2.25x faster and 4.96\% better than LongViT. Notably, while the runtimes are comparable, \name{} is 6.05\% better than MambaVision. Additional experimental results from our 50-epoch runs are available in the supplementary material (\cref{tab:model-comparison-50ep}).

\begin{table}[h]
\centering
\small
\setlength{\tabcolsep}{3pt}
\caption{Comparison of vision backbones on 1024x1024 image resolution on the HR-IN100 benchmark. Each model is evaluated on runtime (in hours), relative speed compared to Atlas, and Top-1 accuracy (in \%). All models are base scale and were trained for 320 epochs until convergence on single 8 $\times$ H100 GPU node.}
\label{tab:model-comparison}
\begin{tabular}{@{}l@{\hspace{2pt}}l@{\hspace{4pt}}c@{\hspace{4pt}}c@{\hspace{4pt}}c@{}}
\toprule
\multicolumn{2}{@{}c}{\textbf{Architecture}} & \textbf{Runtime} & \textbf{Relative} & \textbf{Top-1 Acc.} \\
\multicolumn{2}{@{}l}{} & \textbf{(hr) $\downarrow$} & \textbf{speedup $\downarrow$} & \textbf{(\%) $\uparrow$} \\
\midrule
\multirow{2}{*}{\textbf{Transformer}} & ViT-B & 26.77 & 1.15x & 90.66 \\
& Swin-B & 37.25 & 1.6x & 90.89 \\
& FasterViT-4 & 68.31 & 2.9× & 83.66 \\
& LongViT-B & 52.23 & 2.2× & 86.08 \\
\midrule
\textbf{Convolutional} & ConvNext-B & 100.11 & 4.3× & 91.92 \\
\midrule
\textbf{Mamba} & MambaVision-B & 22.69 & 0.98× & 84.86 \\
\midrule
\textbf{Multi-Scale} & Atlas-B & {\bf 23.12} & {\bf 1.00×} & {\bf 91.04} \\
\bottomrule
\end{tabular}
\end{table}

\begin{table*}[t]
\centering
\small
\setlength{\tabcolsep}{8pt}
\caption{Comparison of Mamba-based (MambaVision-S/16) and Multi-Scale Attention (Atlas-S/16) models across three image resolutions: 1024px, 2048px, and 4096px. The table presents both computational efficiency (runtime in hours on single 8xH100 node) and performance (Top-1 accuracy in \%) metrics. All models were trained for 100 epochs per resolution. Atlas-S/16 demonstrates superior accuracy across all resolutions, with particularly significant advantages at higher resolutions (2048px and 4096px), while maintaining comparable computational demands. The substantial increase in runtime as resolution scales highlights the computational challenges inherent in high-resolution image processing.}
\label{tab:atlas-mamba-4k}
\begin{tabular}{@{}l@{ }l@{ }|ccc|ccc}
\toprule
\multicolumn{2}{c}{\textbf{Model}} & \multicolumn{3}{|c}{\textbf{Runtime (hr)} $\downarrow$} & \multicolumn{3}{|c}{\textbf{Top-1 Accuracy (\%)}  $\uparrow$} \\
&& 1024px  & 2048px & 4096px & 1024px  & 2048px & 4096px \\ 
\midrule
\multirow{1}{*}{\textbf{Mamba-Based}} &
MambaVision-S/16  & 4.56  & 14.73 & 55.5 & 78.2 &  51.42  & 23.36 \\  
\midrule
\multirow{1}{*}{\textbf{Multi-Scale Attention}} &
Atlas-S/16  & 3.64  & 14.23 & 54.72 & \textbf{81.82} & \textbf{67.92} & \textbf{55.84} \\  
\bottomrule
\end{tabular}
\end{table*}

{\bf Long-Context Image Modeling}: The results in~\cref{tab:atlas-mamba-4k} demonstrate the superior scaling capabilities of \name{} over MambaVision on high-resolution images. While both architectures show comparable runtime efficiency on a single 8$\times$H100 node (MambaVision requiring 4.56, 14.73, and 55.5 hours for 1024px, 2048px, and 4096px respectively), Atlas-S/16 outperforms MambaVision-S/16 by 3.62\% at 1024px resolution (81.82\% vs. 78.82\%), with this gap widening to 16.50\% at 2048px and 32.84\% at 4096px. These results highlight \name{}'s capability to effectively capture long-range dependencies at extreme context lengths up to 64K tokens where state-space based models struggle.

\subsection{Ablations}

\noindent{}\textbf{Attention Mechanism.} To understand the efficacy of the MSA block, we run controlled ablations against existing primitives for long-context modelling. The results in \cref{tab:block-comparison} highlight the effectiveness of MSA for classification. While faster in runtime, the window-attention blocks of WViT and Swin perform $\sim$29\% worse than MSA. MambaVisionMixer \cite{hatamizadeh2024mambavision} performs $\sim$12\% worse than MSA while requiring 0.88x the runtime. MSA outperforms the standard attention-block of ViT and the Hierarchichal Attention block from \cite{hatamizadeh2023fastervit}, both in runtime and accuracy. MSA is 1.76$\times$ faster and $\sim$10\% better than ViT block, while being 1.15$\times$ faster and 27$\%$ better than Hierarchical Attention.

\begin{table}[h]
\centering
\small
\setlength{\tabcolsep}{3pt} 
\renewcommand{\arraystretch}{1.2} 
\caption{Comparing different attention mechanisms at a block-level in controlled setting (100epoch runs).\vspace{2mm}}
\label{tab:block-comparison}
\begin{tabular}{l|ccc}
    \toprule
    \textbf{Block} & \textbf{Runtime } & \textbf{Relative} & \textbf{Top-1 acc.} \\
     & \textbf{(min)} $\downarrow$ & \textbf{speedup} $\downarrow$ & \textbf{(in \%)}$\uparrow$ \\
    \midrule
    Window ViT  & 55 & 0.60$\times$ & 41.65 \\
    ShiftedWindow ViT (Swin) & 68 & 0.75$\times$ & 41.48 \\
    ViT & 160 & 1.76$\times$ & 60.57 \\
    \midrule
    Hierarchical Attn. (FasterViT)  & 105 & 1.15$\times$ & 43.19 \\
    Dilated Attn. (LongViT)  & 218 & 2.39$\times$ & 49.88 \\
    MambaVisionMixer & 80 & 0.88$\times$ & 58.79 \\
    {\bf Multi-Scale Attn. (Atlas)} & {\bf 91} & {\bf 1.00$\times$} & {\bf 70.81} \\
    \bottomrule
\end{tabular}
\vspace{-2mm}
\end{table}

Finally, the MSA block is 2.39x faster and 20.9\% better than Dilated Attention block from LongViT \cite{ding2023longnet}. Our results suggest that the MSA block can be used as drop-in replacement to existing primitives, offering significant improvements for long-context modeling.

\noindent{}\textbf{Communication Mechanism.} The MSA block develops a bi-directional communication to efficiently model long-context modeling. The results in \cref{tab:top-down-ablation} demonstrate that MSA significantly improves on vanilla Window-Self Attention (WA), improving classification accuracy by $\sim$12.5\% (72.02 vs 59.39). Furthermore, we show that relying only on multi-scale features with WA is suboptimal, resulting in a 6.9\% performance drop.
While the {\it top-down} and {\it bottom-up} communication mechanisms, independently boost the accuracy by $\sim$3.5\% each, they are complementary to each other. Using the bi-directional communication strategy (i.e. MSA) improves $\sim$3\% over relying on only one of the mechanisms.

\begin{table}[h]
    \centering
    \small
    \setlength{\tabcolsep}{10pt} 
    \renewcommand{\arraystretch}{1.2}
    \vspace{-2.5mm}
    \caption{Ablations on the communication strategies.\vspace{2mm}}
    \label{tab:top-down-ablation}
    \begin{tabular}{l|c}
        \toprule
        \textbf{Communication Strategy} & \textbf{Top-1 (\%)} $\uparrow$ \\
        \midrule 
        single-scale only (WindowViT) & 59.39 \\
        multi-scale only  & 65.14 \\
        multi-scale + bottom-up & 69.92 \\
        multi-scale + top-down  & 69.04 \\
        multi-scale + bottom-up + top-down {\bf (MSA)}  & \textbf{72.02} \\
        \bottomrule
    \end{tabular}
    \vspace{-2.5mm}
\end{table}

\noindent{}\textbf{Composition Strategies.} 
Next, we studied how to best compose MSA blocks into an efficient macro-architecture.
As shown in \cref{tab:composition-comparison}, stacking MSA blocks without progressive downsampling resulted in an accuracy of 69.88\% at a runtime of 75 minutes. Convolutional downsampling between MSA blocks accelerated training, with a runtime of 40 minutes; however, this led to a significant performance drop, with accuracy decreasing to 56.14\%. The Atlas-specific D2D10 configuration, which progressively processes lower-resolution scales, emerged as the most effective strategy, achieving the highest accuracy of 70.09\% at a runtime of 38 minutes. Our novel composition strategy is both led to faster runtimes than traditional convolutional downsampling while yielding comparable performance to no downsampling.

\begin{table}[h]
    \centering
    \small
    \setlength{\tabcolsep}{10pt} 
    \renewcommand{\arraystretch}{1.2}
    \vspace{-2mm}
    \caption{Comparison of different composition strategies.\vspace{2mm}}
    \label{tab:composition-comparison}
    \begin{tabular}{l|cc}
        \toprule
        \textbf{Composition} & \bf \textbf{Runtime (m)} $\downarrow$  & \textbf{Top-1 (\%)} $\uparrow$ \\
        \midrule
        Stack  & 75m  & 69.88 \\
        Downsample (Conv)  & 40m  & 56.14 \\
        {\bf Atlas (D2D10) (ours)} & {\bf 38m}  & \textbf{70.09} \\
        \bottomrule
    \end{tabular}
    \vspace{-2.5mm}
\end{table}

\section{Conclusion}
\label{sec:conclusion}

We propose Multiscale Attention (MSA), a novel primitive for long-context image modeling.
In a controlled block-level experiment, we demonstrated that MSA significantly outperformed alternative cross-token communication strategies, including FasterVIT’s Hierarchical Attention block, and MambaVision Mixer.  MSA achieves this performance through two key insights: multi-scale representations and bidirectional cross-scale communication. 
Building on rich multi-scale representations introduced by MSA, we propose \name, a novel neural network architecture for long context modeling. In system-level experiments, we find that \name{}  significantly improves accuracy-runtime trade-offs in efficient long-context modeling, achieving massive gains over FasterViT, MambaVision, ConvNext, Swin and LongViT. Overall, these results demonstrate that multi-scale attention significantly improves long-context image modeling. 

\section{Acknowledgements} We thank the UCSF Facility of Advanced Computing team, including Hunter McCallum, Sandeep Giri, Rhett Hillary, Marissa Jules, Sean Locke, and John Gallias, for their work in supporting our computational environment. This was supported by a grant from EvansMDS, a funding initiative of the Edward P. Evans Foundation. Research reported in this publication was also supported by the National Cancer Institute of the National Institutes of Health under Award Number R37CA289821. The content is solely the responsibility of the authors and does not necessarily represent the official views of the National Institutes of Health.

\bibliography{references}
\bibliographystyle{icml2025}

\clearpage
\appendix

\section{Implementation Details}
\label{sec:implementation_details}
For the baselines that we compared with in~\cref{tab:model-comparison}, we utilize the provided code as is, without modifications to gradient accumulation, employing a linearly decaying learning rate proportional to the batch size, following~\cite{goyal2017accurate}. This ensures consistency with prior work and facilitates a fair comparison of performance. For the various model hyperparamters, we use the configs as provided by authors for Imagenet-1K where available.

\section{Additional Experiments}
\label{sec:additional_experiments}

\begin{table*}[htbp]
    \centering
    \small
    \setlength{\tabcolsep}{8pt}
    \caption{Comparison of vision models across different image resolutions. Each model has two rows: one for runtime (in minutes) and one for Top-1 accuracy (in \%). We trained all models for 50 epochs for each resolution. We limited each experiment to a maximum runtime of 24hrs on an 8 $\times$ H100 GPU node and report “--” for experiments that could not be complete within our runtime limit.}
    \label{tab:model-comparison-50ep}
    \begin{tabular}{@{}l@{ }l@{ }|cccc|cccc}
    \toprule
    \multicolumn{2}{c}{\textbf{Model}} & \multicolumn{4}{|c}{\textbf{Runtime (min)} $\downarrow$} & \multicolumn{4}{|c}{\textbf{Top-1 Accuracy (\%)}  $\uparrow$} \\
    && 256px & 512px & 1024px & 2048px & 256px & 512px & 1024px & 2048px \\ 
    \midrule
    \multirow{2}{*}{\textbf{Transformer-Based}} &
    ViT-B/16  & 18 & 51 & 247 & 3480 & 63.68 & 72.60 & 69.42 & -- \\ 
    & WViT-B/16 & 18 & 44 & 137 & 638 & 64.21 & 68.95 & 63.61 & 53.93 \\ 
    \midrule
    \multirow{1}{*}{\textbf{Convolutional}} &
    ConvNext-B/16 & 66 & 237 & 955 & 3825 & 78.84 & 75.94 & 67.50 & -- \\  
    \midrule
    \multirow{2}{*}{\textbf{Sparse-Transformer}} &
    FasterViT-4 & 49 & 168 & 675 & 2400  & 77.64 & 74.40 & 53.62 & -- \\ 
    & LongViT-B/16 & 39 & 116 & 442 & 2000  & 55.20 & 51.88 & 45.32 & -- \\  
    \midrule
    \multirow{1}{*}{\textbf{Mamba-Based}} &
    MambaVision-B/16  & 21 & 56 & 197 & 750 & 73.10 & 69.94 & 51.68 & 24.64 \\  
    \midrule
    \multirow{1}{*}{\textbf{Multi-Scale Attention}} &
    Atlas-B/16  & 25 & 54 & 198 & 786 & \textbf{80.05} & \textbf{83.75} & \textbf{82.73} & \textbf{74.74} \\  
    \bottomrule
    \end{tabular}
    \end{table*}

To validate our findings, we conducted 50-epoch training runs following prior work showing that shorter training schedules still provide reliable signals about architectural performance ~\cite{xiao2021early}. These abbreviated runs maintain the same relative performance trends across architectures while requiring significantly less computational resources. As shown in Table 1, Atlas-B/16 maintains its superior accuracy-runtime trade-off across resolutions, achieving high accuracy while maintaining reasonable training times even at 2048px resolution, where several competing architectures exceed our 24-hour runtime limit.

\section{Additional Optimizations: QKV Caching for Multi-Scale Attention}
\label{sec:cache}
A naive implementation of Multi-Scale Attention (MSA) would require recomputing Query, Key, and Value (QKV) projections for each window involved in cross-attention operations across different scales.  Consider a window $W^{(l)}$ at scale $l$ performing cross-attention with windows at coarser scales $\{W^{(l+1)}, \ldots, W^{(L)}\}$ in the top-down pathway.  In a naive implementation, the QKV for each window $W^{(l)}$ would be recalculated for every cross-attention instance, even if the underlying feature representation of $W^{(l)}$ remains unchanged.  This repeated computation becomes increasingly inefficient as the number of scales and windows grows.

To overcome this challenge, we introduce a QKV cache mechanism within MSA.  During both the top-down and bottom-up pathways, we maintain a cache at each scale $l$ to store the QKV projections for all windows $\{W^{(l)}_{ij}\}$.  When a window at scale $l$ needs to perform cross-attention, it first queries this cache. If a valid QKV set for the current version of $W^{(l)}$ is available, it is directly retrieved from the cache. The cache is updated only when the feature representation of a window at a given scale is modified. This occurs after self-attention at the coarsest scale $L$, and after each dense cross-attention operation in the top-down and parent cross-attention in the bottom-up pathways. By reusing QKV projections, our cache signficantly accelerates MSA in long sequences where cross-scale attention operations are frequent.

\end{document}